\documentclass{article}

\usepackage{arxiv}

\usepackage[utf8]{inputenc} 
\usepackage[T1]{fontenc}    
\usepackage{hyperref}       
\usepackage{url}            
\usepackage{booktabs}       
\usepackage{amsfonts}       
\usepackage{nicefrac}       
\usepackage{microtype}      
\usepackage{lipsum}		
\usepackage{graphicx}
\usepackage[numbers]{natbib}
\usepackage{doi}
\usepackage{xcolor}
\usepackage{amsmath}
\usepackage[]{threeparttable}

\title{Benchmarking Vision, Language, \& Action Models on Robotic Learning Tasks
}

\date{} 					

\author{ Pranav Guruprasad\thanks{equal contribution, alphabetical order. Corresponding Author: \texttt{pranav@metarch.ai} \ \textbf{Sponsored by Metarch.ai}}
 \ $^1$$^2$, Harshvardhan Sikka$^*$$^1$$^2$$^3$, Jaewoo Song$^*$$^1$, Yangyue Wang$^1$, Paul Pu Liang$^4$
}

\date{%
    $^1$Manifold Research\\%
    $^2$Metarch.ai\\%
    $^3$Georgia Tech\\
    $^4$MIT
    }

\renewcommand{\headeright}{}
\renewcommand{\undertitle}{}

\begin{document}
\maketitle

\begin{abstract}
Vision-language-action (VLA) models represent a promising direction for developing general-purpose robotic systems, demonstrating the ability to combine visual understanding, language comprehension, and action generation. However, systematic evaluation of these models across diverse robotic tasks remains limited. In this work, we present a comprehensive evaluation framework and benchmark suite for assessing VLA models. We profile three state-of-the-art VLM and VLAs —GPT-4o, OpenVLA, and JAT—across 20 diverse datasets from the Open-X-Embodiment collection, evaluating their performance on various manipulation tasks. Our analysis reveals several key insights: (1) current VLA models show significant variation in performance across different tasks and robot platforms, with GPT-4o demonstrating the most consistent performance through sophisticated prompt engineering, (2) all models struggle with complex manipulation tasks requiring multi-step planning, and (3) model performance is notably sensitive to action space characteristics and environmental factors. We release our evaluation framework and findings to facilitate systematic assessment of future VLA models and identify critical areas for improvement in the development of general-purpose robotic systems.
\end{abstract}

\begin{figure}[h]
	\centering
	\includegraphics[width=15cm]{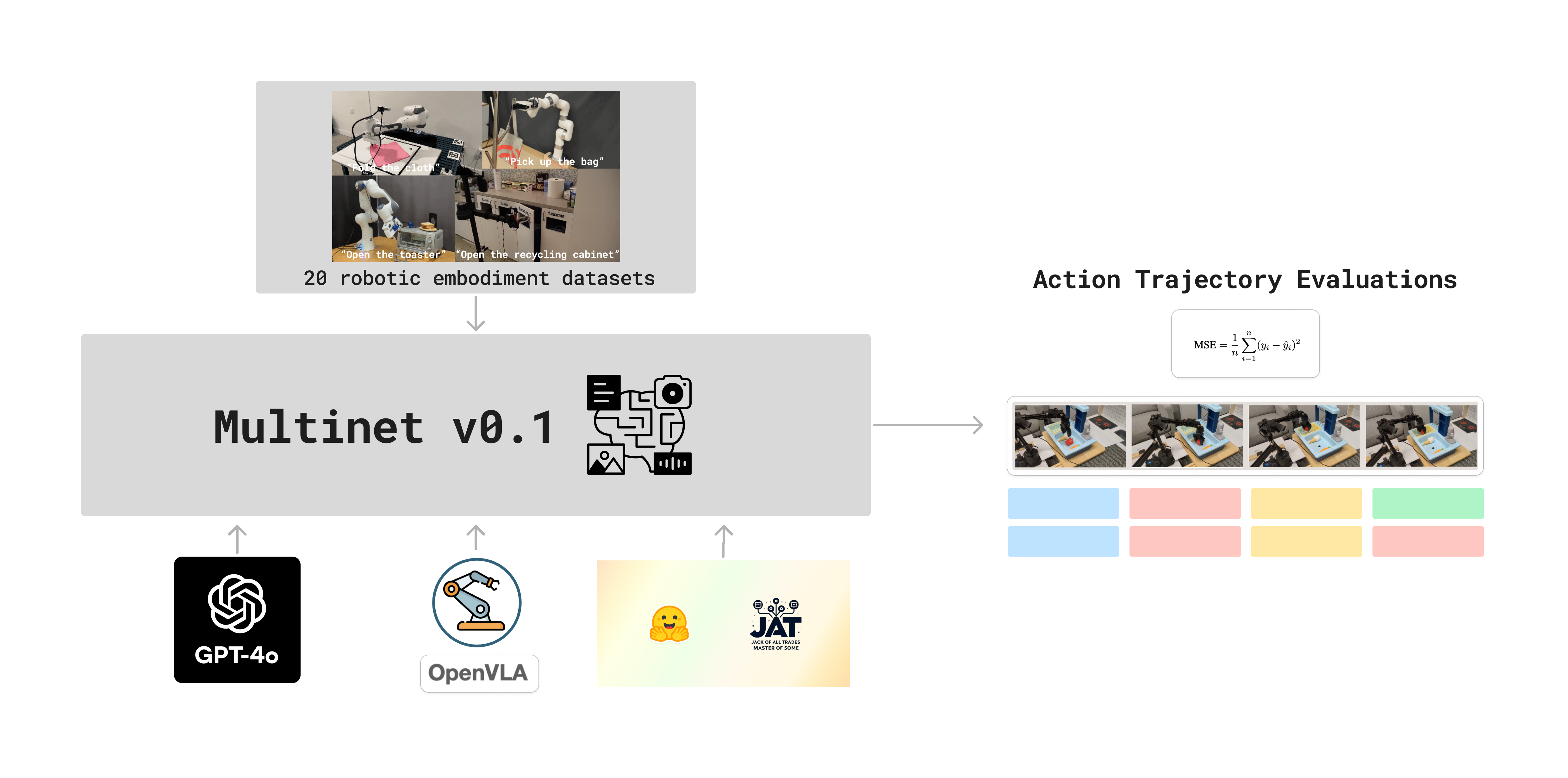}
	\caption{Multinet v0.1 overview. Benchmarking a SoTA VLM, SoTA VLA, and novel generalist model on 20 real-world robotics datasets by comparing the action predicted by the models, with the ground truth action from the dataset, at every timestep.}
	\label{fig:fig1}
\end{figure}


\section{Introduction}

The quest for robust, generalizable robotic systems continues to pose a fundamental challenge in machine learning and robotics research. Despite significant progress in controlled environments, current systems exhibit limited generalization beyond their training conditions. These limitations span numerous dimensions: systems fail when encountering unfamiliar task descriptions \cite{brohan2023rt, walke2023bridgedata}, struggle with spatial variations in object configurations \cite{brohan2022rt}, perform poorly under variable lighting or occlusion \cite{chi2023diffusion}, and show degraded performance when interacting with novel objects or in cluttered environments \cite{xie2024decomposing,team2024octo}. These generalization challenges significantly hinder the deployment of learned robotic systems in unconstrained environments.

Recent breakthroughs in foundation models, especially in vision and language processing, suggest a promising path forward. These models, trained on web-scale datasets, have achieved remarkable capabilities in visual understanding \cite{kirillov2023segment,minderer2022simpleopenvocabularyobjectdetection}, sophisticated reasoning about interactions between objects and agents \cite{alayrac2024multi, driess2023palm, wang2022git}, software development \cite{chen2021evaluating}, and cross-modal comprehension. The robust generalization exhibited by these models addresses precisely the challenges that have historically limited robotics systems. Their advanced capabilities in semantic understanding, problem-solving, and visual processing could revolutionize the development of versatile robots capable of handling diverse tasks in dynamic environments.

This approach corresponds with a broader trend in machine learning toward unified neural sequence architectures. These models demonstrate continued performance gains at the boundaries of data volume, computational resources, and model complexity \cite{kaplan2020scaling,hoffmann2022training}. This pattern aligns with historical observations suggesting that general-purpose models efficiently utilizing computational resources tend to outperform specialized solutions \cite{sutton2019bitter}. The advantages of unified sequence models are multifaceted: they remove the requirement for custom policy architectures with domain-specific assumptions, enable the use of diverse training data through sequence-based representation, and show reliable improvements with increasing scale.

Nevertheless, adapting these models for robotics applications presents substantial challenges. The vast scale of training data available for foundation models - billions of tokens and images from the internet - far exceeds what is currently feasible to collect for robot interactions \cite{open_x_embodiment_rt_x_2023,khazatsky2024droid}. Moreover, while foundation models excel at abstract reasoning and high-level comprehension, robotic control requires precise, physically grounded actions, such as specific end-effector movements. Recent research has explored integrating language models (LLMs) and vision-language models (VLMs) into robotics frameworks \cite{ahn2022icanisay,driess2023palm,vemprala2023chatgptroboticsdesignprinciples}.  However, many current approaches limit foundation models to high-level planning roles, using them essentially as advanced state machines that convert commands into basic actions, executed by separate low-level controllers unable to access the models' rich semantic understanding.

Current research initiatives have investigated leveraging pretrained language and vision-language models to enhance robotic representations \cite{shridhar2022cliport, nair2022r3m, karamcheti2023language}. These components have also been integrated into planning systems \cite{driess2023palm, stone2023open}. A particularly promising development has been the emergence of vision-language-action models (VLAs), which extend foundation models for robotics through various approaches including pretraining \cite{brohan2022rt,team2024octo} or fine-tuning \cite{kim2024openvla, brohan2023rt, marcu2024lingoqavisualquestionanswering}. These models have shown encouraging results in transferring to novel tasks, marking an important advancement toward developing generally capable robotic systems.

As these models continue to evolve, there is a critical need for systematic evaluation of their capabilities across both their intended multimodal training domains and out-of-distribution scenarios. 

Our primary contributions in this paper are:
\begin{itemize}
  \item Detailed profiling results for an initial set of VLM, VLA, and emerging ``generalist'' models, providing insights into their capabilities and limitations.
  \item Analysis of model generalization to a diverse set of real-world robotics datasets comprising a wide variety of tasks and environments.
  \item A systematic set of evaluation splits and metrics specifically designed for robotics learning tasks in the widely-used OpenX Dataset.
  \item A general framework for mapping VLMs to other modality classes, with particular emphasis on action spaces.
  \item Open-source software infrastructure for downloading, managing, and utilizing the benchmark data.
\end{itemize}

Through this work, we aim to provide the robotics learning community with robust tools and methodologies for assessing and comparing these emerging approaches, facilitating progress in this rapidly evolving field and helping to bridge the gap between foundation models and practical robotics applications. Importantly, this is the first foray into a new large-scale generalist action model benchmark, called MultiNet v0.1, which we discuss in the context of Future Work.

\section{Related Work}
Recent years have seen a proliferation of benchmarks aimed at evaluating multimodal models across different domains and capabilities. We organize our discussion of related work into three categories: general multimodal benchmarks, robotics-specific benchmarks, and multimodal language model evaluations.

\paragraph{General Multimodal Benchmarks}
MultiBench \cite{liang2021multibench} represents one of the first systematic attempts to evaluate multimodal learning across diverse domains, spanning healthcare, robotics, affective computing, and finance. Similar to our work, MultiBench emphasizes the importance of evaluating multiple aspects of model performance, including generalization, complexity, and robustness. However, while MultiBench covers a broad range of domains, its robotics evaluation is limited in scope. MMMU \cite{yue2024mmmu} provides another comprehensive benchmark focused on college-level multimodal understanding. The authors evaluate models across technical disciplines like engineering and science through expert-level problems requiring nuanced perception and domain-specific knowledge, but do not specifically address robotics control tasks.

\paragraph{Multimodal Language Model Evaluations} The evolution of multimodal evaluation has progressed from single-task benchmarks like VQA \cite{antol2015vqa}, OK-VQA \cite{marino2019ok}, MSCOCO \cite{lin2014microsoft}, and GQA \cite{hudson2019gqa} to more comprehensive evaluation frameworks. Recent benchmarks span various capabilities, from basic OCR to adversarial robustness and hallucination detection (e.g., POPE \cite{li2023evaluating} and HaELM \cite{wang2023evaluation}). More holistic evaluations have emerged through benchmarks like LAMM \cite{yin2023lamm}, LVLM-eHub \cite{xu2023lvlm}, SEED \cite{ge2023planting}, MMBench \cite{zhang2023mmbench}, and MM-Vet \cite{yu2023mm}. Specialized benchmarks such as MathVista \cite{lu2023mathvista} focus on specific domains like mathematical reasoning, while GAIA \cite{mialon2023gaia} tests fundamental abilities in reasoning and multimodality handling.

\paragraph{Robotics-Specific Benchmarks} The evolution of robotics datasets has demonstrated considerable diversity across various dimensions, particularly with the advancement of imitation learning and behavior cloning (BC). While many robotics benchmarks focus on evaluating model adaptability to new tasks, functionalities, or environments, there remains a gap in systematically evaluating different BC models at scale in both simulated and real-world settings. THE COLOSSEUM \cite{pumacay2024colosseum} addresses this gap by providing a systematic evaluation framework focused on robotic manipulation, evaluating generalization across 14 different environmental perturbations. Similar efforts include FactorWorld \cite{xie2024decomposing}, which examines 11 variation factors across 19 tasks, and KitchenShift \cite{xing2021kitchenshift}, which evaluates zero-shot generalization across 7 variation factors in kitchen environments.Several other specialized robotics benchmarks have emerged: RLBench \cite{james2019rlbench} offers a suite of 100 manipulation tasks in simulation; RAVENS \cite{huang2023raven} focuses on vision-based manipulation; and FurnitureBench \cite{heo2023furniturebench} provides reproducible real-world benchmarks for long-horizon complex manipulation. LIBERO \cite{liu2024libero} offers benchmarks for knowledge transfer in lifelong robot learning, while FMB \cite{luo2023fmb} emphasizes generalizable robotic learning across complex tasks. Recent work has also introduced DUDE \cite{van2023document} for robotic document manipulation and ProcTHOR \cite{deitke2022️} for large-scale embodied AI using procedural generation.

Our work differs from these previous benchmarks in several key aspects. First, we focus specifically on evaluating models' ability to process and generate actions from real-world robotic trajectories, rather than simulated environments or static vision-language tasks. Second, by leveraging the OpenX dataset, we evaluate across a diverse range of robot platforms and tasks, providing a more comprehensive view of model capabilities. Third, our evaluation framework specifically measures models' ability to perform zero-shot generalization across different action spaces and robot morphologies, a crucial capability for general-purpose robotic systems.

\section{Evaluating VLMs and VLAs}
\subsection{Data}

Our evaluation framework leverages the Open X-Embodiment Dataset (OpenX), currently the largest open-source repository of real robot trajectories. OpenX represents a significant collaborative effort across 21 institutions, aggregating over 1 million real robot trajectories from 22 distinct robot embodiments, ranging from single-arm manipulators to bi-manual systems and quadrupedal robots. The dataset's comprehensive nature makes it particularly suitable for evaluating generalist models, as it spans a diverse range of manipulation and locomotion tasks, environmental conditions, and robot configurations.

The dataset utilizes the Reinforcement Learning Datasets (RLDS) format, storing data in serialized tfrecord files. This standardized format efficiently accommodates the heterogeneous nature of robotics data, handling varied action spaces and input modalities across different robot setups. For instance, the format seamlessly integrates data from systems with different sensor configurations, including varying numbers of RGB cameras, depth sensors, and point cloud generators.

For version 0.1 of our benchmark, we utilize 53 of the 72 available OpenX datasets, as detailed in Table \ref{tab:all-datasets}. We present results for 20 of these datasets for all 3 models, and have the full 53 for JAT.This subset was selected to ensure comprehensive coverage across different task types, embodiments, and environmental conditions while maintaining data quality and consistency. For datasets that did not include pre-defined evaluation sets, we have created and provided new evaluation splits to ensure robust assessment of model performance. The training splits of these 53 datasets comprise approximately 32 terabytes of data.

This careful curation of the OpenX dataset provides several advantages for our evaluation framework:

\begin{enumerate}
\item Scale and Diversity: The large number of trajectories and varied robot embodiments allows for comprehensive assessment of model generalization capabilities.
\item Real-World Relevance: Being composed entirely of real robot data rather than simulated interactions, the dataset better reflects the challenges of physical robot deployment.
\item Standardization: The consistent RLDS format facilitates systematic evaluation across different robot platforms and task types.
\item Cross-Domain Assessment: The inclusion of both manipulation and locomotion tasks enables evaluation of model performance across fundamentally different types of robot control.
\end{enumerate}

The complete list of included datasets and their characteristics is provided in the appendix.

\subsubsection{Dataset Curation}

To ensure the quality and utility of our benchmark, we implemented a systematic curation process for the OpenX datasets. This process was designed to maximize the diversity and relevance of the included data while maintaining practical considerations for large-scale evaluation.

Our curation methodology consisted of several steps. First, we conducted a high-level review of dataset quality and accessibility, which resulted in the exclusion of three datasets: Austin BUDS, Austin Sailor, and Stanford Kuka Multimodal. For datasets that contained only training splits, we performed a detailed comparative analysis based on the robot platform used for data collection. This analysis considered multiple features: Robot model and morphology, Gripper specifications, Action space characteristics, Sensor configuration (number and type of RGB cameras, depth cameras, and wrist-mounted cameras), Presence of language annotations, Availability of camera calibration data, and Inclusion of proprioceptive information.
  
When multiple datasets shared identical values across all these features for the same robot platform, we retained only the dataset with the larger number of episodes. This decision was made to minimize redundancy while maximizing the diversity of our evaluation set. This approach ensures that each included dataset contributes unique information to the benchmark, either through different robot configurations, sensor setups, or task specifications.

Several additional datasets were excluded from version 0.1 of our benchmark due to technical limitations in their accessibility through the TensorFlow Datasets (TFDS) builder, which is the recommended data loading mechanism for OpenX. These compatibility issues will be addressed in future versions of the benchmark as the underlying infrastructure evolves.
This careful curation process results in a benchmark that balances comprehensive coverage with practical considerations, ensuring that the included datasets provide meaningful evaluation scenarios while maintaining manageable computational requirements. 

\subsection{Models}

In our evaluation, we focus on three recent vision-language-action (VLA) models that represent the current state-of-the-art in generalist robot learning: JAT (Jack of All Trades), GPT-4o, and OpenVLA. These models are particularly noteworthy for their ability to handle multiple modalities and their demonstrated capabilities across a wide variety of tasks.

JAT \cite{gallouédec2024jacktradesmastersome} is a transformer-based model optimized for handling sequential decision-making tasks and multi-modal data types. With 768-dimensional hidden states and 12 layers, JAT employs a dual attention mechanism inspired by the Longformer architecture, combining global attention with a 512-token window and local attention with a 256-token window. The model was trained for 250,000 steps on a diverse dataset spanning robotics control, computer vision, and natural language processing tasks. JAT's architecture is specifically designed to provide wider attention windows for timesteps compared to previous approaches, making it particularly suitable for long-horizon robotics tasks.

GPT-4o \cite{achiam2023gpt} represents a significant advancement in omni-modal modeling, accepting combinations of text, audio, image, and video inputs while generating multi-modal outputs. The model demonstrates strong performance in robotic manipulation tasks, particularly in scenarios requiring generalization to novel objects and environments. GPT-4o incorporates advanced safety measures and has been extensively evaluated across multiple risk categories, including cybersecurity, persuasion, and model autonomy.

OpenVLA, a 7B-parameter open-source vision-language-action model, was trained on 970,000 robot episodes from the Open X-Embodiment dataset. Its architecture combines a 600M-parameter visual encoder (utilizing both SigLIP and DinoV2 models) with a 7B-parameter Llama 2 language model backbone. OpenVLA is notable for its strong performance in generalist robot manipulation tasks, outperforming larger models while using significantly fewer parameters. The model particularly excels in multi-task environments involving multiple objects and demonstrates strong language grounding abilities.

Each of these models represents different approaches to the challenge of generalist robot learning:

JAT emphasizes broad "generalist" multi-modal capabilities.
GPT-4o is a powerful VLM, and allows for various approaches to map language output to action \& control tasks.
OpenVLA prioritizes open-source accessibility while maintaining competitive performance with larger closed-source models

This diversity in approaches provides valuable insights into different architectural and training strategies for generalist robot learning. The models also represent different points on the spectrum of model size and computational requirements, allowing us to evaluate the relationship between model scale and performance across various robotics tasks.

\subsection{Evaluation Metrics}

Mean Squared Error (MSE) serves as our primary metric for evaluating model performance on offline robotics trajectories. In the context of offline reinforcement learning, MSE has proven to be a reliable metric for estimating optimal value functions and has demonstrated strong empirical performance. For our benchmark, MSE is particularly appropriate due to several key properties:

\begin{enumerate}
   \item Non-Negativity: The metric remains non-negative, ensuring that errors are consistently accounted for without potential cancellation effects from opposing signs.
   \item Sensitivity to Large Errors: The squared term in MSE emphasizes larger deviations, providing clear indication of significant prediction errors.
   \item Bias-Variance Trade-off: MSE inherently captures both bias and variance components, offering a comprehensive measure of prediction accuracy.
\end{enumerate}

For a given prediction, MSE is calculated as:

\begin{equation}
   \text{MSE} = \frac{1}{n} \sum_{i=1}^{n} (y_i - \hat{y}_i)^2
\end{equation}

where $y_i$ represents the ground truth action, $\hat{y}_i$ is the predicted action, and $n$ is the number of observations.

For our benchmark, we employ MSE to evaluate how accurately models predict actions given the observation states, image observation, and language instruction at each timestep. Given the offline nature of the OpenX dataset and the inability to evaluate models on physical robots, comparing predicted and ground truth action tensors provides the most direct assessment of model performance.

We report several variations of MSE to provide a comprehensive performance analysis:

\begin{enumerate}
   \item \textbf{Average MSE (AMSE):} Computed as the mean MSE across all timesteps in a dataset, AMSE enables direct comparison of model performance across different datasets and architectures.

   \item \textbf{Normalized AMSE (NAMSE):} Calculated as an average of the min-max normalized MSEs over all the timesteps in the dataset, this metric normalizes prediction errors to each model's error range, facilitating more equitable cross-dataset for a single model comparison by accounting for different scales in model outputs and dataset action spaces.

   \item \textbf{Completion Rate:} We assess successful completion by comparing final predicted actions with ground truth final actions for all episodes in the dataset. While this serves as an approximate measure of task completion, it provides valuable insights into models' ability to reach target states across trajectories.
\end{enumerate}

The combination of these metrics allows us to evaluate both the fine-grained accuracy of action predictions and the overall task-completion capabilities of different models. This is particularly important in offline robotics, where environments and rewards are not available for policy evaluation.

\section{Experimental Setup}

\subsection{Profiling Configuration}

We established specific configurations for each model to ensure consistent and fair evaluation across the diverse OpenX datasets. Below, we detail the precise setup for each model, including handling of inputs, processing decisions, and any necessary adaptations.

\paragraph{JAT Configuration}
The JAT model was evaluated in a zero-shot setting, where predictions are made using only the current timestep information without access to previous states. For each prediction, the model receives the observation state, observation image, and language instruction. Several key preprocessing steps were implemented:

\begin{itemize}
    \item \textbf{Image Processing:} JAT requires 4-channel images. For 3-channel RGB inputs, we create an RGBA image by duplicating the red channel as the alpha channel. For 2-channel inputs, we duplicate both channels to create a 4-channel representation.
    \item \textbf{Observation Processing:} For dictionary-type observations, we concatenate all floating-point observations (excluding image and language instruction embeddings) into a single tensor. In cases where no floating-point observations exist, we pass a zero-filled dummy tensor.
    \item \textbf{Action Processing:} Ground truth actions are processed by concatenating all floating-point actions into a single tensor when the action space is represented as a dictionary.
    \item \textbf{Multi-Image Handling:} For timesteps with multiple available images, we select the primary image (typically designated with the keyword `image').
\end{itemize}

\paragraph{GPT Configuration}
GPT was also evaluated in a zero-shot configuration, with several specific processing requirements:

\begin{itemize}
    \item \textbf{Prompt Construction:} Each prediction is based on a comprehensive prompt including:
    \begin{itemize}
        \item Floating-point observation states with their corresponding keys as descriptors for specific datasets like Berkeley Autolab where there are such observation states available.
        \item Primary image observation
        \item Natural language instruction
        \item Verbal descriptions for each action space dimension
        \item The official action space statistics if available or statistical information (min, max, mean) for each action dimension.
        \item Environmental and task descriptions when available
    \end{itemize}
    \item \textbf{Output Processing:} To handle GPT's VLM-native outputs, which may be incompatible with the required floating-point action tensor format, we implemented error handling:
    \begin{itemize}
        \item For incompatible outputs (incorrect tensor sizes, string elements, mixed text-tensor outputs, or non-scalar elements), we generate a random action tensor with values in $[0.0, 1.0)$ as a fallback.
    \end{itemize}
    \item \textbf{Multi-Image Processing:} For timesteps with multiple available images, we select the primary image (typically designated with the keyword `image').
\end{itemize}

\paragraph{OpenVLA Configuration}
OpenVLA's configuration focused primarily on action space handling and gripper command conversions:

\begin{itemize}
    \item \textbf{Gripper Command Standardization:} We implemented several conversion protocols:
    \begin{itemize}
        \item Binary discretization: For $[0,1]$ to $\{0,1\}$ conversion, we threshold at $0.5$
        \item Ternary discretization: For $[0,1]$ to $\{-1,0,1\}$ conversion, values $<0.05$ map to $-1$ (closed), $>0.95$ to $1$ (open), and $[0.05,0.95]$ to $0$ (no change)
        \item Continuous normalization: For $[0,1]$ to $[-1,1]$ conversion, we apply the formula: $y = 2 \cdot (x - orig_{low}) / (orig_{high} - orig_{low}) - 1$. This was used by the authors in \citet{kim2024openvla}.
    \end{itemize}
    \item \textbf{Special Cases:}
    \begin{itemize}
        \item For the UCSD pick-and-place dataset, we used dataset statistics to scale gripper commands to the appropriate torque space
        \item For ETH agent affordances, we applied the transformation: $unnormalized = 0.5 \cdot (normalized + 1) \cdot (high - low) + low$, where high and low are the 99th and 1st percentiles respectively
    \end{itemize}
    \item \textbf{Action Space Handling:}
    \begin{itemize}
        \item For datasets using velocity, angular velocity, or torque-based action spaces (e.g., ETH agent affordances and UCSD datasets), we note potential compatibility issues with OpenVLA's position-based predictions
        \item We exclude `Terminal' tensors from action spaces, as OpenVLA predicts only XYZ, RPY, and gripper commands
    \end{itemize}
\end{itemize}

\paragraph{Additional Considerations}
We encountered cases where image observations were unavailable due to non-standard image key naming (e.g., `agentview\_rgb', `frontright\_fisheye\_image') in some datasets. These were utilized for OpenVLA, but not the other models, as OpenVLA requires an image as part of its input. This specific case occurred with 2 datasets  in particular,  conq\_hose\_manipulation, and viola.

\subsection{Inference Infrastructure}

To facilitate reproducible evaluation of these models, we detail the infrastructure requirements and setup for each model's inference pipeline.

\paragraph{JAT and GPT Infrastructure}
For JAT evaluation and GPT API interfacing, we utilized a Google Cloud Platform (GCP) e2-standard-8 instance with 8 vCPU (4 physical cores), 32 GB memory, and x86/64 architecture. While this configuration exceeds the minimum requirements, the additional computational resources enabled efficient parallelization of evaluation runs. For GPT specifically, as inference occurs through API endpoints, the local infrastructure requirements are minimal. Storage was provided through GCP's standard persistent disk service.

\paragraph{OpenVLA Infrastructure}
OpenVLA inference was conducted on a GCP g2-standard-8 instance equipped with a single NVIDIA L4 GPU, 8 vCPU (4 physical cores), 32 GB system memory, and x86/64 architecture. The NVIDIA L4 GPU, featuring the Ada Lovelace architecture, was specifically chosen for two key advantages: compatibility with Flash Attention 2.x for efficient attention computation, and 24 GB of GDDR6 memory, sufficient for full-model inference of OpenVLA without optimization. Storage was similarly provided through GCP's standard persistent disk service.

\section{Results \& Discussion}

In our evaluation of vision-language-action models, we seek to answer several fundamental questions about their capabilities and limitations: (1) How do current VLM \& VLA models perform across diverse robotics tasks and platforms, particularly in zero-shot settings? (2) What impact do different model architectures and training approaches (e.g., prompt engineering, robotics-specific training) have on performance? (3) How well do these models handle different action spaces and robot morphologies? (4) What are the current limitations and failure modes of these models in real-world robotics tasks? Through systematic analysis of three state-of-the-art models across 20 diverse datasets, we provide insights into these questions below.

\begin{figure}[h]
	\centering
	\includegraphics[width=15cm]{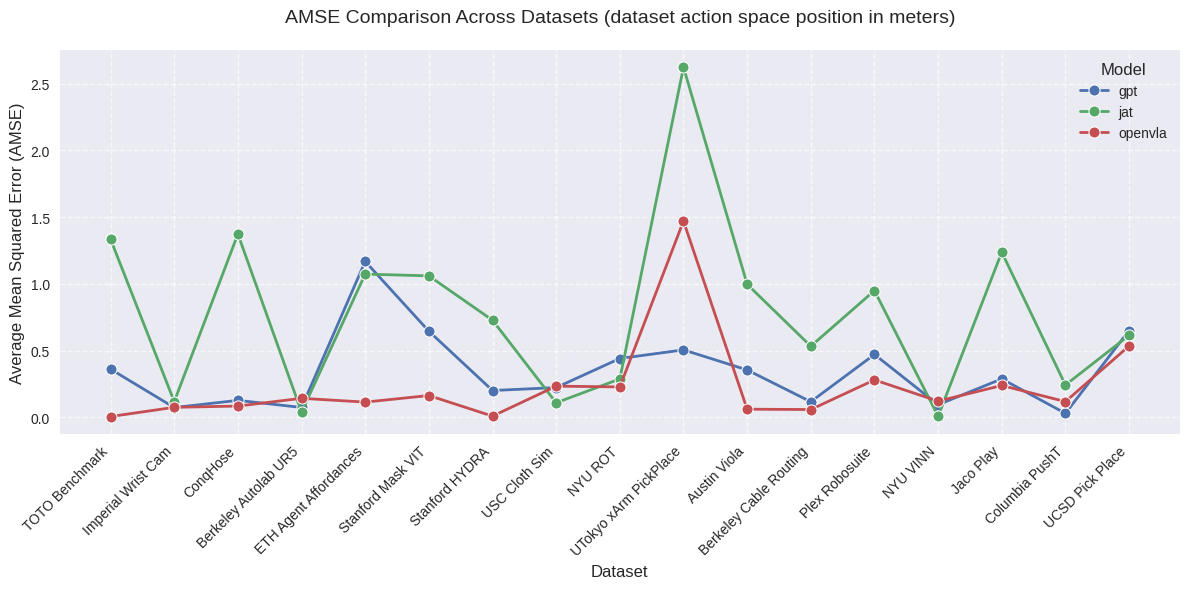}
	\caption{AMSE values of GPT-4o, JAT, and OpenVLA Across 20 OpenX Datasets. JAT displays the poorest performance out of the 3 models with higher AMSE scores, while OpenVLA and GPT-4o demonstrate similar performance. OpenVLA displays consistent performance across most datasets}
	\label{fig:fig0}
\end{figure}

\begin{table*}[t]
\centering
\setlength{\tabcolsep}{4pt}  
\caption{Dataset Coverage and Action Space Characteristics}
\label{tab:dataset-coverage}
\resizebox{1.0\linewidth}{!}{   
\begin{tabular}{llll}
\toprule
Dataset Name & Registered Dataset Name & In Pretraining OpenVLA & Action Space Type \\
\midrule
Jaco Play & jaco\_play & \checkmark & 4D (1 grip, 3 pos) \\
Berkeley Cable Routing & berkeley\_cable\_routing & \checkmark & 7D (3 ang, 3 pos, 1 term) \\
NYU Door Opening & nyu\_door\_opening\_surprising\_effectiveness & & 8D (1 grip, 3 ang, 3 pos, 1 term) \\
VIOLA & viola & \checkmark & 8D (1 grip, 3 ang, 3 pos, 1 term) \\
Berkeley Autolab UR5 & berkeley\_autolab\_ur5 & \checkmark & 8D (1 grip, 3 ang, 3 pos, 1 term) \\
TOTO & toto & \checkmark & 7D (3 ang, 3 pos, 1 term) \\
Columbia PushT & columbia\_cairlab\_pusht\_real & & 8D (1 grip, 3 ang, 3 pos, 1 term) \\
NYU ROT & nyu\_rot\_dataset\_converted\_externally\_to\_rlds & & 7D (3 pos, 3 ang, 1 grip) \\
Stanford HYDRA & stanford\_hydra\_dataset\_converted\_externally\_to\_rlds & \checkmark & 7D (3 pos, 3 ang, 1 grip) \\
UCSD Kitchen & ucsd\_kitchen\_dataset\_converted\_externally\_to\_rlds & \checkmark & 8D (3 pos, 3 ang, 1 grip, 1 term) \\
UCSD Pick Place & ucsd\_pick\_and\_place\_dataset\_converted\_externally\_to\_rlds & & 4D (3 vel, 1 grip torque)\\
USC Cloth Sim & usc\_cloth\_sim\_converted\_externally\_to\_rlds & & 4D (3 pos, 1 grip)  \\
Tokyo PR2 Fridge & utokyo\_pr2\_opening\_fridge\_converted\_externally\_to\_rlds &  & 8D (3 pos, 3 ang, 1 grip, 1 term) \\
Tokyo PR2 Tabletop & utokyo\_pr2\_tabletop\_manipulation\_converted\_externally\_to\_rlds &  & 8D (3 pos, 3 ang, 1 grip, 1 term)  \\
UTokyo xArm Pick-Place & utokyo\_xarm\_pick\_and\_place\_converted\_externally\_to\_rlds &  & 7D (3 pos, 3 ang, 1 grip) \\
Stanford MaskVIT & stanford\_mask\_vit\_converted\_externally\_to\_rlds & & 5D (3 pos, 1 ang, 1 grip)  \\
ETH Agent Affordances & eth\_agent\_affordances & & 6D (3 vel, 3 ang vel)  \\
Imperial Sawyer & imperialcollege\_sawyer\_wrist\_cam & & 8D (3 pos, 3 ang, 1 grip, 1 term) \\
ConqHose & conq\_hose\_manipulation & & 7D (3 pos, 3 ang, 1 grip)  \\
Plex RoboSuite & plex\_robosuite & & 7D (3 pos, 3 ang, 1 grip) \\
\bottomrule
\end{tabular}}
\begin{tablenotes}
\centering
\small
\item pos: position, orient: orientation, grip: gripper, term: terminate, vel: velocity, ang: angular
\end{tablenotes}
\end{table*}

\begin{table*}[t]
\centering
\caption{Performance Metrics Comparison across Models}
\label{tab:performance-comparison}
\resizebox{0.8\linewidth}{!}{
\begin{tabular}{lcccccc}
\toprule
& \multicolumn{2}{c}{GPT} & \multicolumn{2}{c}{OpenVLA} & \multicolumn{2}{c}{JAT} \\
\cmidrule(lr){2-3} \cmidrule(lr){4-5} \cmidrule(lr){6-7}
Dataset Name & AMSE & NAMSE & AMSE & NAMSE & AMSE & NAMSE \\
\midrule
Jaco Play & 0.288 & 0.188 & 0.239 & 0.228 & 1.237 & 0.295 \\
Berkeley Cable Routing & 0.117 & 0.010 & 0.058 & 0.091 & 0.533 & 0.411 \\
NYU Door Opening & 0.094 & 0.046 & 0.121 & 0.304 & 0.008 & 0.061 \\
VIOLA & 0.355 & 0.134 & 0.061 & 0.072 & 0.997 & 0.331 \\
Berkeley Autolab UR5 & 0.074 & 0.049 & 0.142 & 0.249 & 0.040 & 0.073 \\
TOTO & 0.361 & 0.069 & 0.006 & 0.004 & 1.335 & 0.238 \\
Columbia PushT & 0.030 & 0.046 & 0.118 & 0.820 & 0.242 & 0.347 \\
NYU ROT & 0.441 & 0.034 & 0.228 & 0.308 & 0.288 & 0.177 \\
Stanford HYDRA & 0.201 & 0.009 & 0.009 & 0.054 & 0.728 & 0.147 \\
UCSD Kitchen & 11580.963 & 0.207 & 5018.936 & 0.116 & 34890.635 & 0.353 \\
UCSD Pick Place & 0.650 & 0.086 & 0.535 & 0.175 & 0.614 & 0.210 \\
USC Cloth Sim & 0.223 & 0.260 & 0.234 & 0.305 & 0.109 & 0.375 \\
Tokyo PR2 Fridge & 16035.136 & 0.037 & 68433.175 & 0.159 & 221666.531 & 0.324 \\
Tokyo PR2 Tabletop & 2550.878 & 0.014 & 8728.959 & 0.116 & 117663.493 & 0.364 \\
UTokyo xArm Pick-Place & 0.505 & 0.088 & 1.471 & 0.252 & 2.623 & 0.254 \\
Stanford MaskVIT & 0.645 & 0.120 & 0.163 & 0.184 & 1.060 & 0.571 \\
ETH Agent Affordances & 1.168 & 0.057 & 0.114 & 0.139 & 1.073 & 0.290 \\
Imperial Sawyer & 0.073 & 0.183 & 0.075 & 0.517 & 0.118 & 0.356 \\
ConqHose & 0.127 & 0.024 & 0.084 & 0.264 & 1.373 & 0.178 \\
Plex RoboSuite & 0.471 & 0.067 & 0.280 & 0.206 & 0.950 & 0.142 \\
\bottomrule
\end{tabular}}
\begin{tablenotes}
\centering
\small
\item AMSE: Average Mean Squared Error, NAMSE: Normalized Average Mean Squared Error
\item Large AMSE values (e.g., for Kitchen and PR2 tasks) reflect different action space scales
\end{tablenotes}
\end{table*}

\subsection{Average Model Performance Analysis}

Our evaluation reveals significant variations in performance across models and datasets. We observe that while JAT consistently shows higher AMSE (indicating worse performance) across most datasets, OpenVLA and GPT demonstrate more comparable performance levels, with AMSE typically below 0.5 for most datasets.

\paragraph{Overall Performance Patterns}
For OpenVLA, we observe generally consistent performance across most datasets with AMSE in the 0.1-0.5 range, with best performance of all 3 models for tasks that fall within its training distribution, with notable exceptions in complex manipulation tasks. GPT shows comparable or slightly better performance on many datasets, particularly excelling in precise manipulation tasks. Both models maintain relatively stable performance across similar task types, though with different error profiles.

GPT demonstrates strongest performance on:
\begin{itemize}
   \item berkeley\_autolab\_ur5 (AMSE: 0.074)
   \item columbia\_cairlab\_pusht\_real (AMSE: 0.030)
   \item imperialcollege\_sawyer\_wrist\_cam (AMSE: 0.073)
\end{itemize}

\paragraph{Common Challenges}
Both models exhibit significant challenges with certain task types:
\begin{itemize}
   \item Complex manipulation tasks, particularly those involving large movements or multi-step sequences like Kitchen manipulation tasks. 
   \item Tasks requiring significant temporal reasoning or complex action sequences. This follows naturally as the models were assessed in a zero shot fashion. 
\end{itemize}

\subsubsection{Model-Specific Analysis}

The performance patterns we observe can may be attributable to several architectural and training differences between the models:

\paragraph{OpenVLA}
The combination of SigLIP and DinoV2 visual encoders appears to provide robust visual features, contributing to consistent performance across tasks. However, this comes at the cost of absolute precision in some cases. The model's specific training on robotics data from OpenX likely contributes to its stability across different task types, though it may not always achieve optimal performance on any single task type.

\paragraph{GPT}
GPT's sophisticated prompt construction and ability to handle detailed statistical information about action spaces appears to help in making more precise predictions for well-defined tasks. Its strong performance on precise manipulation tasks suggests that its general-purpose capabilities transfer well to robotics control in structured scenarios. However, it shows similar limitations to OpenVLA in complex, multi-step tasks.

\paragraph{JAT}
JAT's significantly higher AMSE across datasets suggests that its architecture, while suitable for general-purpose tasks, may not be optimized for precise robotics control. 

\subsubsection{Implications for Future Development}

These results suggest several directions for improvement in VLA model development:
\begin{itemize}
   \item The variation in performance across robot platforms suggests that more work is needed in developing platform-agnostic control capabilities
   \item The superior performance of GPT and OpenVLA in their respective strengths suggests that combining their approaches - sophisticated prompt engineering with robotics-specific training - might yield better overall performance
\end{itemize}

\begin{figure}[h]
	\centering
	\includegraphics[width=16cm]{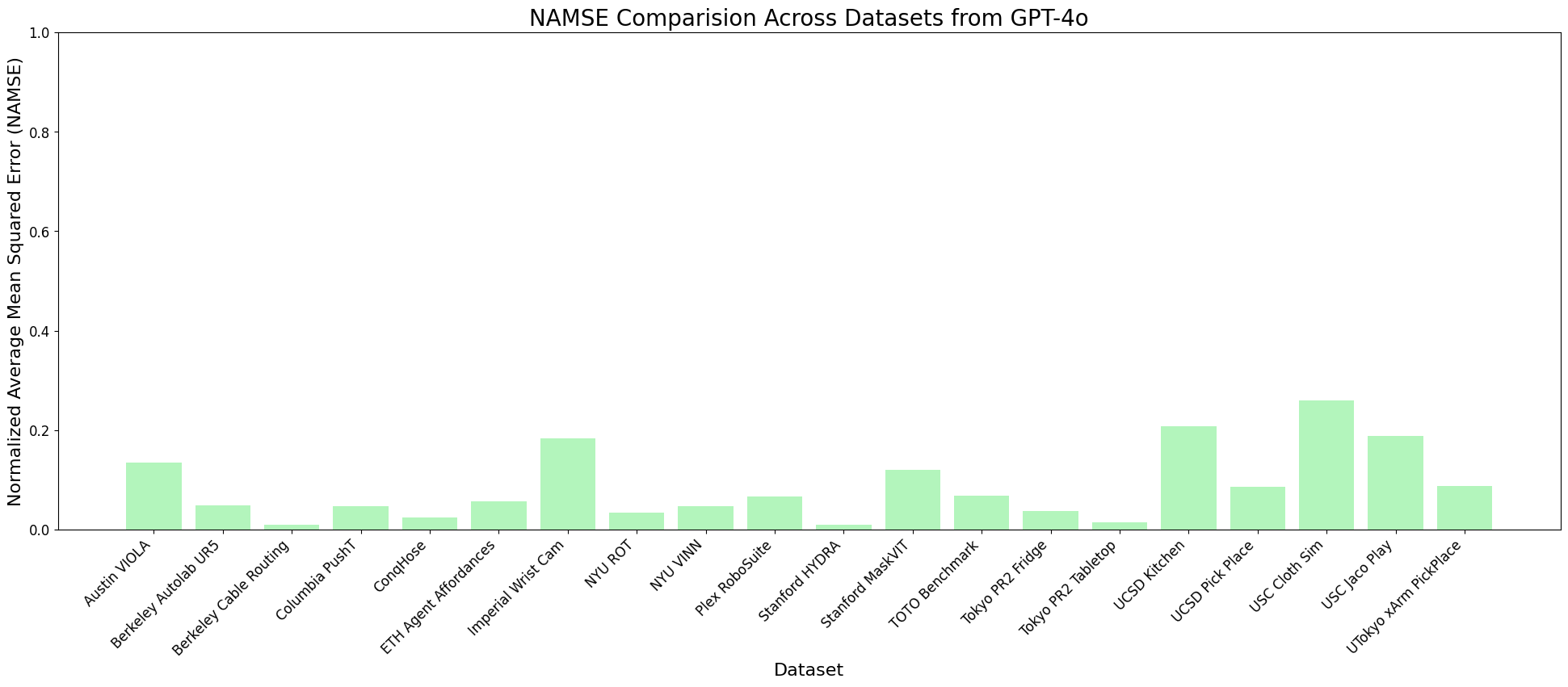}
	\caption{Normalized AMSE For GPT4o. GPT-4o demonstrates consistent NAMSE across all datasets, suggesting that the prompt engineering framework which provides detailed information about the action space, task, and environment, may be a key factor.}
	\label{fig:fig2}
\end{figure}

\begin{figure}[h]
	\centering
	\includegraphics[width=16cm]{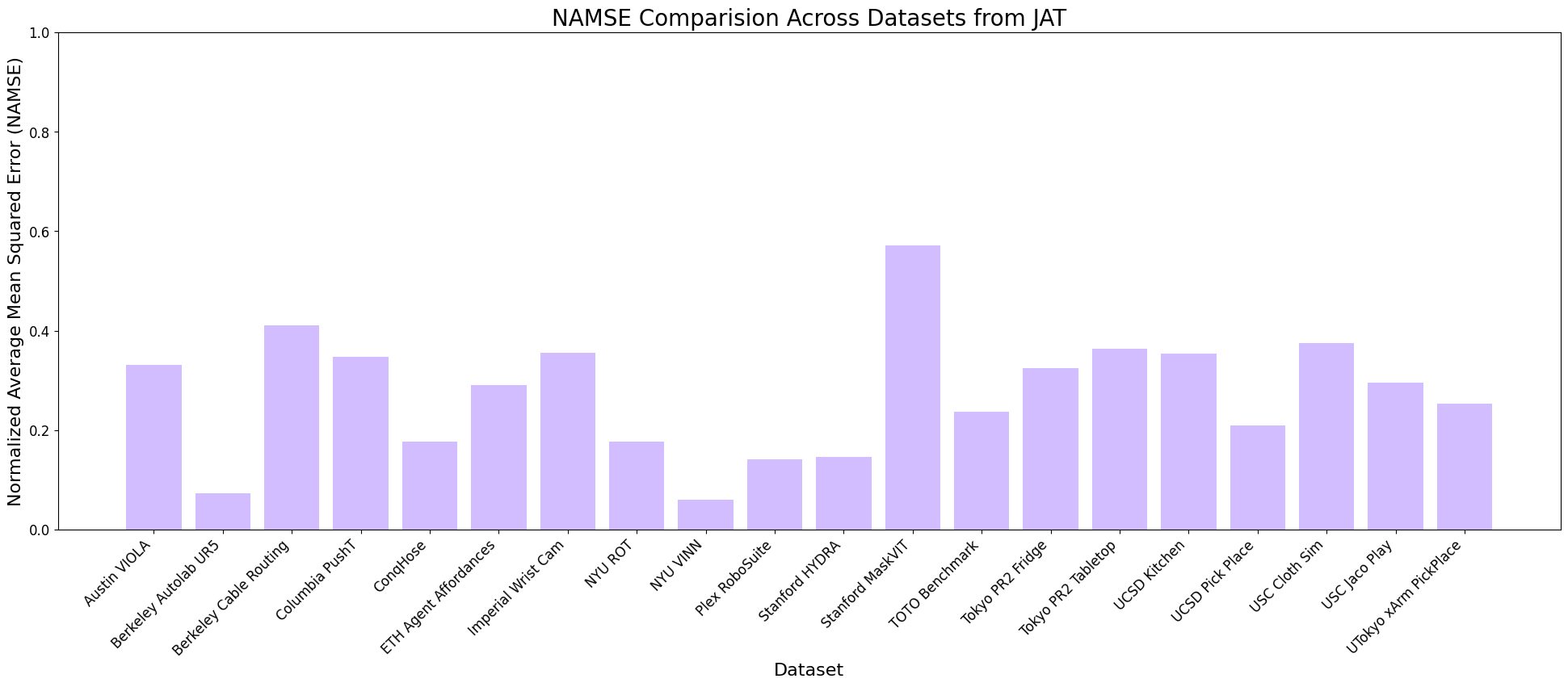}
	\caption{Normalized AMSE For JAT. JAT exhibits moderate variation in NAMSE across tasks, with a spike in the Stanford MaskVIT Dataset, while maintaining realtively consistent preformance for similar task types.}
	\label{fig:fig3}
\end{figure}

\begin{figure}[h]
	\centering
	\includegraphics[width=16cm]{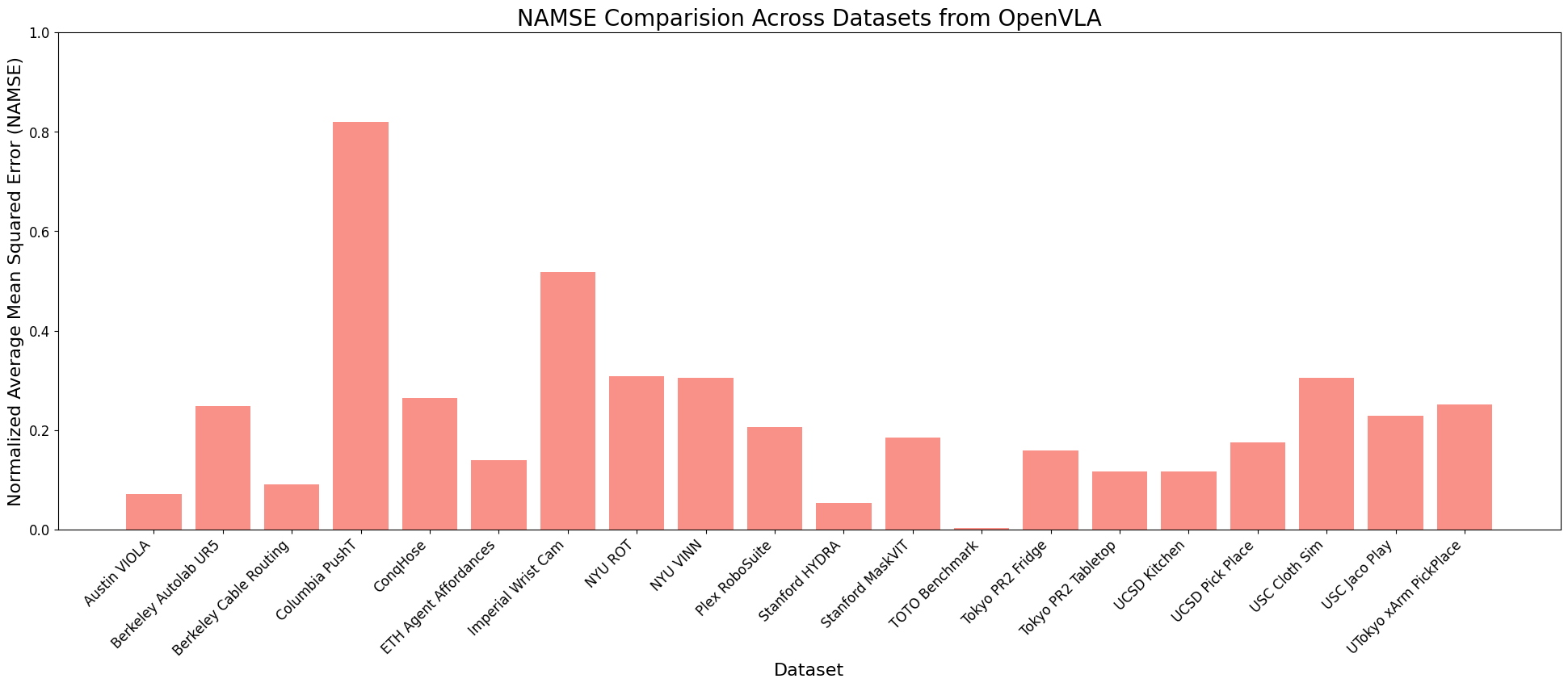}
	\caption{Normalized AMSE For OpenVLA. OpenVLA shows high variation in NAMSE values. As expected it displays very strong performance on tasks such as toto which is within its training distribution. OpenVLA shows a clear pattern of task-specific performance differences.}
	\label{fig:fig4}
\end{figure}

\subsection{Normalized Performance Analysis}

While absolute performance metrics like AMSE provide insight into task-specific capabilities, normalized average mean squared error (NAMSE) allows us to understand how each model performs across different tasks relative to its own capabilities. NAMSE is particularly valuable for understanding inherent task difficulty and model behavior patterns independent of action space scale.

\subsubsection{Model-Specific Performance Patterns}

\paragraph{GPT-4o}
GPT-4o demonstrates remarkably consistent normalized performance across datasets, with NAMSE generally remaining below 0.2. This stability is particularly noteworthy given the diversity of tasks in the benchmark. The model's sophisticated prompt engineering approach appears to be a key factor in this consistency, as it includes:
\begin{itemize}
   \item Explicit action space statistics (min, max, mean) for each dimension
   \item Verbal descriptions for each action dimension
   \item Detailed environment and task descriptions when available
\end{itemize}
This comprehensive prompting strategy provides clear constraints and context for each prediction, likely contributing to the model's ability to maintain consistent relative performance across diverse tasks.

\paragraph{OpenVLA}
OpenVLA shows the most dramatic variation in normalized performance:
\begin{itemize}
   \item Highest normalized error on columbia\_cairlab\_pusht\_real (NAMSE: 0.82)
   \item Exceptionally strong performance on pretrained tasks (e.g., toto with NAMSE: 0.003)
   \item Clear pattern of task-specific performance variations
\end{itemize}
This variation suggests that OpenVLA's architecture and training approach may lead to stronger task specialization compared to other models.

\paragraph{JAT}
JAT exhibits moderate variation across tasks, with NAMSE typically ranging from 0.2 to 0.4:
\begin{itemize}
   \item Notable performance spike on stanford\_mask\_vit (NAMSE $\sim$0.57)
   \item Relatively consistent performance band for similar task types
   \item Higher baseline NAMSE compared to GPT-4o but more stable than OpenVLA
\end{itemize}

\subsubsection{Cross-Model Insights}

The normalized analysis reveals several key patterns about task difficulty and model architecture:

\paragraph{Task Difficulty Patterns}
Certain tasks consistently show higher normalized error across all models, independent of architecture:
\begin{itemize}
   \item Complex manipulation tasks and multi-step operations consistently show higher NAMSE
   \item Simple pick-and-place operations tend to show lower normalized error
   \item Tasks requiring precise control generally result in higher normalized error
\end{itemize}

\paragraph{Architectural Implications}
The variation in normalized performance across models provides insights into their architectural strengths:
\begin{itemize}
   \item GPT-4o's consistent normalized performance suggests its architecture and prompting strategy create a more generally robust system
   \item OpenVLA's high variation indicates stronger task specialization, possibly due to its training approach and dual visual encoder
   \item JAT's moderate but consistent variation suggests a middle ground between specialization and generalization
\end{itemize}

This normalized analysis reveals that while absolute performance varies significantly, there are consistent patterns in what tasks are relatively more challenging for each model architecture. The success of GPT-4o's prompt engineering approach, in particular, suggests that providing structured context about action spaces and environmental constraints may be a key factor in achieving consistent performance across diverse tasks. This observation could inform future development of VLA models, suggesting that incorporating more explicit task and action space information could improve robustness and generalization capabilities.

\subsection{Key Takeaways}

Our evaluation of VLM \& VLA models reveals several fundamental insights about the current state of VLM \& VLA models transferring to robotics tasks:

1. Prompt Engineering Impact: GPT-4o's consistent performance across diverse tasks demonstrates that structured prompting with explicit action space information and environmental context may be as important as architectural choices. This suggests that future VLA development should consider incorporating structured task representations as a core design principle.

2. Specialization vs. Generalization: We observe a clear trade-off between specialized and general performance. OpenVLA shows superior performance on tasks within its training distribution but higher variation across tasks, while GPT-4o maintains more consistent but sometimes suboptimal performance. This highlights the ongoing challenge of developing models that can both specialize and generalize effectively.

3. Task Complexity Barriers: All models, regardless of architecture, struggle with complex manipulation tasks requiring multi-step planning or precise control. This consistent limitation suggests that current approaches may be missing key capabilities needed for complex robotics tasks.

4. Action Space Sensitivity: Performance varies significantly with different action space characteristics, particularly in tasks requiring precise control or complex action sequences. This suggests the need for more robust methods of handling diverse action spaces and robot morphologies.

\section{Future Work}

While our current results provide valuable insights into the capabilities and limitations of these models, we envision several important directions for expanding and enhancing this benchmark. We present these as a subset of a larger benchmark we are developing, dubbed MultiNet. We contextualize the opportunities ahead in the context of this benchmark below.

A critical question in the development of generalist models is whether the integration of control capabilities comes at the cost of performance in other domains. To address this, future versions of MultiNet will evaluate SOTA VLAs on pure vision-language and language tasks, allowing us to assess whether fine-tuning or co-training on control tasks impacts their performance in these foundational modalities. This analysis will help inform architectural and training strategies that maintain strong performance across all modalities.

We also plan to expand beyond the OpenX dataset to evaluate these models on a broader range of control tasks. This expansion will allow us to better understand how VLAs and generalist models perform on completely out-of-distribution data, providing insights into their true generalization capabilities. While our current evaluations focus on zero-shot performance, future work will investigate few-shot learning and fine-tuning scenarios, offering a more complete picture of these models' adaptability.

A particularly promising direction is the exploration of VLA transfer to non-robotic domains. We are especially interested in investigating how these models can be fine-tuned for software environments, potentially enabling the development of more capable digital agents. This research could reveal insights about the generalization of embodied learning principles to virtual environments.

Additionally, we identify several novel directions for future investigation:

\begin{itemize}
   \item \textbf{Compositional Generalization:} Evaluating how well VLAs can combine learned primitives to solve novel tasks, particularly in scenarios requiring multi-step reasoning or tool use.
   
   \item \textbf{Long Sequence Reliability:} Developing metrics to assess the consistency of model behavior over extended sequences, including the ability to maintain goals and adapt to changing conditions.
   
   \item \textbf{Cross-Embodiment Transfer:} Further investigating how knowledge transfers between different robot morphologies, potentially leading to more efficient training strategies for new platforms.
   
   \item \textbf{Memory and Long-Term Planning:} Assessing models' capabilities in tasks requiring long-term memory and strategic planning, particularly in multi-phase manipulation tasks.
   
   \item \textbf{Multi-Agent Interaction:} Extending the benchmark to scenarios involving multiple agents, evaluating coordination and collaborative manipulation capabilities.
\end{itemize}

Finally, while MultiNet currently operates as an offline benchmark, we plan to develop online evaluation capabilities. This expansion will include the integration of simulation environments for both 2D and 3D control tasks, enabling more dynamic and interactive assessment of model performance. Such environments will allow for more comprehensive evaluation of model capabilities in real-time decision-making scenarios.

Through these future developments, we aim to establish MultiNet as a comprehensive and rigorous benchmark for assessing and advancing the field of vision-language-action models. This expanded scope will provide researchers and practitioners with valuable tools for understanding and improving these increasingly important models.

\section{Conclusion}

In this work, we presented a comprehensive evaluation framework for vision-language-action models and conducted a systematic analysis of their performance across a diverse range of robotics tasks. Our study reveals several important insights about the current state of VLA models and highlights critical areas for future development.

We find that current VLA models demonstrate varying levels of capability across different tasks, with notable strengths and limitations. GPT-4o shows remarkable consistency in normalized performance across datasets, likely due to its sophisticated prompt engineering approach that provides structured context about action spaces and environmental constraints. OpenVLA demonstrates strong performance on certain tasks but shows higher variation across different scenarios, suggesting task-specific specialization. JAT, while showing moderate consistency, generally achieves higher error rates, indicating potential limitations in its architecture for precise control tasks.

Our analysis reveals several critical challenges that need to be addressed in future work. First, all models struggle significantly with complex manipulation tasks. Second, the performance of these models varies substantially across different robot platforms and action spaces, suggesting a need for more robust architectures that can better handle diverse control scenarios. Third, the notable impact of prompt engineering on performance, as demonstrated by GPT-4o, suggests that developing more sophisticated ways to provide context and constraints to these models could be a promising direction for improvement.

Looking forward, our results suggest several promising directions for future research. The development of more robust architectures that can maintain consistent performance across diverse tasks while handling complex, multi-step manipulations remains a key challenge. Additionally, the integration of structured task representations and better handling of temporal dependencies could help address the current limitations in complex manipulation tasks. Finally, our open-source evaluation framework provides a foundation for systematic assessment of future VLA models, enabling more rigorous comparison and benchmarking of new approaches. We are excited to engage with the broader research community to extend these results and advance the emerging class of Multimodal VLA models.

\bibliographystyle{plainnat}
\bibliography{references}

\section{Appendix} 

\subsection{Dataset Coverage, Completion Rate, and Additional AMSE Recordings}

\begin{table*}[h]
\centering
\caption{Dataset Coverage and Action Space Types}
\label{tab:all-datasets}
\resizebox{1.05\linewidth}{!}{
\begin{tabular}{llcccc}
\toprule
Dataset Name & Registered Dataset Name & JAT & GPT & OpenVLA & Action Space Type \\
\midrule
RT-1 Robot Action & fractal20220817\_data & \checkmark & & & 10D (2 pos for base, 1 ang for base, 1 grip, 3 ang for arm, 3 pos for arm) \\
QT-Opt & kuka & \checkmark & & & 10D (2 pos for base, 1 ang for base, 1 grip, 3 ang for arm, 3 pos for arm) \\
Berkeley Bridge & bridge & \checkmark & & & 7D (3 pos, 3 ang,1 term) \\
Freiburg Franka Play & taco\_play & \checkmark & & & -- \\
USC Jaco Play & jaco\_play & \checkmark & \checkmark & \checkmark & 4D (1 grip, 3 pos) \\
Berkeley Cable Routing & berkeley\_cable\_routing & \checkmark & \checkmark & \checkmark & 7D (3 ang, 3 pos, 1 term) \\
Roboturk & roboturk & \checkmark & & & -- \\
NYU VINN & nyu\_door\_opening\_surprising\_effectiveness & \checkmark & \checkmark & \checkmark & 8D (1 grip, 3 ang, 3 pos, 1 term) \\
Austin VIOLA & viola & \checkmark & \checkmark & \checkmark & 8D (1 grip, 3 ang, 3 pos, 1 term) \\
Berkeley Autolab UR5 & berkeley\_autolab\_ur5 & \checkmark & \checkmark & \checkmark & 8D (1 grip, 3 ang, 3 pos, 1 term) \\
TOTO Benchmark & toto & \checkmark & \checkmark & \checkmark & 7D (3 ang, 3 pos, 1 term)\\
Language Table & language\_table & \checkmark & & & 2D \\
Columbia PushT & columbia\_cairlab\_pusht\_real & \checkmark & \checkmark & \checkmark & 8D (1 grip, 3 ang, 3 pos, 1 term) \\
NYU ROT & nyu\_rot\_dataset\_converted\_externally\_to\_rlds & \checkmark & \checkmark & \checkmark & 7D (3 pos, 3 ang, 1 grip) \\
Stanford HYDRA & stanford\_hydra\_dataset\_converted\_externally\_to\_rlds & \checkmark & \checkmark & \checkmark & 7D (3 pos, 3 ang, 1 grip) \\
NYU Franka Play & nyu\_franka\_play\_dataset\_converted\_externally\_to\_rlds & \checkmark & & & -- \\
Maniskill & maniskill\_dataset\_converted\_externally\_to\_rlds & \checkmark & & & -- \\
Furniture Bench & furniture\_bench\_dataset\_converted\_externally\_to\_rlds & \checkmark & & & 8D (3 pos, 4 quat, 1 grip) \\
CMU Franka Exploration & cmu\_franka\_exploration\_dataset\_converted\_externally\_to\_rlds & \checkmark & & & -- \\
UCSD Kitchen & ucsd\_kitchen\_dataset\_converted\_externally\_to\_rlds & \checkmark & \checkmark & \checkmark & 8D (3 pos, 3 ang, 1 grip, 1 term) \\
UCSD Pick Place & ucsd\_pick\_and\_place\_dataset\_converted\_externally\_to\_rlds & \checkmark & \checkmark & \checkmark & 4D (3 vel, 1 grip torque) \\
Austin Sirius & austin\_sirius\_dataset\_converted\_externally\_to\_rlds & \checkmark & & & -- \\
BC-Z & bc\_z & \checkmark & & & 61D (30 pos, 30 ang, 1 grip) \\
USC Cloth Sim & usc\_cloth\_sim\_converted\_externally\_to\_rlds & \checkmark & \checkmark & \checkmark & 4D (3 pos, 1 grip) \\
Tokyo PR2 Fridge & utokyo\_pr2\_opening\_fridge\_converted\_externally\_to\_rlds & \checkmark & \checkmark & \checkmark & 8D (3 pos, 3 ang, 1 grip, 1 term) \\
Tokyo PR2 Tabletop & utokyo\_pr2\_tabletop\_manipulation\_converted\_externally\_to\_rlds & \checkmark & \checkmark & \checkmark & 8D (3 pos, 3 ang, 1 grip, 1 term) \\
Saytap & utokyo\_saytap\_converted\_externally\_to\_rlds & \checkmark & & & -- \\
UTokyo xArm PickPlace & utokyo\_xarm\_pick\_and\_place\_converted\_externally\_to\_rlds & \checkmark & \checkmark & \checkmark & 7D (3 pos, 3 ang, 1 grip) \\
UTokyo xArm Bimanual & utokyo\_xarm\_bimanual\_converted\_externally\_to\_rlds & \checkmark & \checkmark & & 14D (dual arm 7D control) \\
Berkeley MVP Data & berkeley\_mvp\_converted\_externally\_to\_rlds & \checkmark & & & -- \\
Berkeley RPT Data & berkeley\_rpt\_converted\_externally\_to\_rlds & \checkmark & & & -- \\
KAIST Nonprehensile & kaist\_nonprehensile\_converted\_externally\_to\_rlds & \checkmark & \checkmark & & 20D (3 pos, 3 ang, 7 gain coeff, 7 damping ratio coeff) \\
Stanford MaskVIT & stanford\_mask\_vit\_converted\_externally\_to\_rlds & \checkmark & \checkmark & \checkmark & 5D (3 pos, 1 ang, 1 grip) \\
LSMO Dataset & tokyo\_u\_lsmo\_converted\_externally\_to\_rlds & \checkmark & & & -- \\
ConqHose & conq\_hose\_manipulation & \checkmark & \checkmark & \checkmark & 7D (3 pos, 3 ang, 1 grip) \\
ETH Agent Affordances & eth\_agent\_affordances & \checkmark & \checkmark & \checkmark & 6D (3 vel, 3 ang vel) \\
Imperial Wrist Cam & imperialcollege\_sawyer\_wrist\_cam & \checkmark & \checkmark & \checkmark & 8D (3 pos, 3 ang, 1 grip, 1 term) \\
Plex RoboSuite & plex\_robosuite & \checkmark & \checkmark & \checkmark & 7D (6 pose, 1 grip) \\
DLR Sara Grid Clamp Dataset & dlr\_sara\_grid\_clamp\_converted\_externally\_to\_rlds & \checkmark & & & -- \\
DLR Sara Pour Dataset & dlr\_sara\_pour\_converted\_externally\_to\_rlds & \checkmark & & & -- \\
DLR Wheelchair Shared Control & dlr\_edan\_shared\_control\_converted\_externally\_to\_rlds & \checkmark & & & -- \\
ASU TableTop Manipulationl & asu\_table\_top\_converted\_externally\_to\_rlds & \checkmark & & & -- \\
CMU Franka Pick-Insert Data & iamlab\_cmu\_pickup\_insert\_converted\_externally\_to\_rlds & \checkmark & & & -- \\
Austin Mutex & utaustin\_mutex & \checkmark & & & -- \\
Stanford Robocook & stanford\_robocook\_converted\_externally\_to\_rlds & \checkmark & & & -- \\
CMU Play Fusion & cmu\_play\_fusion & \checkmark & & & -- \\
CMU Stretch & cmu\_stretch & \checkmark & & & -- \\
RECON & berkeley\_gnm\_recon & \checkmark & & & -- \\
CoryHall & berkeley\_gnm\_cory\_hall & \checkmark & & & -- \\
SACSoN & berkeley\_gnm\_sac\_son & \checkmark & & & -- \\
DobbE & dobbe & \checkmark & & & -- \\
IO-AI Office PicknPlace & io\_ai\_tech & \checkmark & & & -- \\
RoboSet & robo\_set & \checkmark & & & -- \\
\bottomrule
\end{tabular}}
\begin{tablenotes}
\centering
\small
\item pos: position, orient: orientation, grip: gripper, term: terminate, vel: velocity, ang: angular, quat: quaternion
\item Some datasets have been excluded due to space constraints or incomplete information
\end{tablenotes}
\end{table*}

\begin{table*}[ht]
\centering
\caption{Task Completion Rates Across Models and Datasets}
\label{tab:completion-rates}
\resizebox{0.5\linewidth}{!}{
\begin{tabular}{lccc}
\toprule
Dataset Name & GPT & OpenVLA & JAT \\
\midrule
Jaco Play & 0.917\% & 29.358\% & 0.000\% \\
Berkeley Cable Routing & 0.000\% & 0.000\% & 0.000\% \\
NYU Door Opening & 0.000\% & 0.000\% & 0.000\% \\
VIOLA & 0.000\% & 0.000\% & 0.000\% \\
Berkeley Autolab UR5 & 1.923\% & 0.000\% & 0.000\% \\
TOTO & 0.000\% & 0.000\% & 0.000\% \\
Columbia PushT & 0.000\% & 0.000\% & 0.000\% \\
NYU ROT & 7.143\% & 0.000\% & 0.000\% \\
Stanford HYDRA & 0.833\% & 0.000\% & 0.000\% \\
UCSD Kitchen & 0.000\% & 0.000\% & 0.000\% \\
UCSD Pick Place & 0.000\% & 0.000\% & 0.000\% \\
USC Cloth Sim & 0.000\% & 0.000\% & 0.000\% \\
Tokyo PR2 Fridge & 0.000\% & 0.000\% & 0.000\% \\
Tokyo PR2 Tabletop & 0.000\% & 0.000\% & 0.000\% \\
UTokyo xArm Pick-Place & 0.000\% & 0.000\% & 0.000\% \\
Stanford MaskVIT & 0.000\% & 0.000\% & 0.000\% \\
ETH Agent Affordances & 0.000\% & 0.000\% & 0.000\% \\
Imperial Sawyer & 0.000\% & 0.000\% & 0.000\% \\
ConqHose & 0.000\% & 0.000\% & 0.000\% \\
Plex RoboSuite & 0.000\% & 0.000\% & 0.000\% \\
\bottomrule
\end{tabular}}
\begin{tablenotes}
\centering
\small
\item Success rates reported as percentage of episodes where final action matched ground truth
\end{tablenotes}
\end{table*}

\begin{figure}[ht]
	\centering
	\includegraphics[width=15cm]{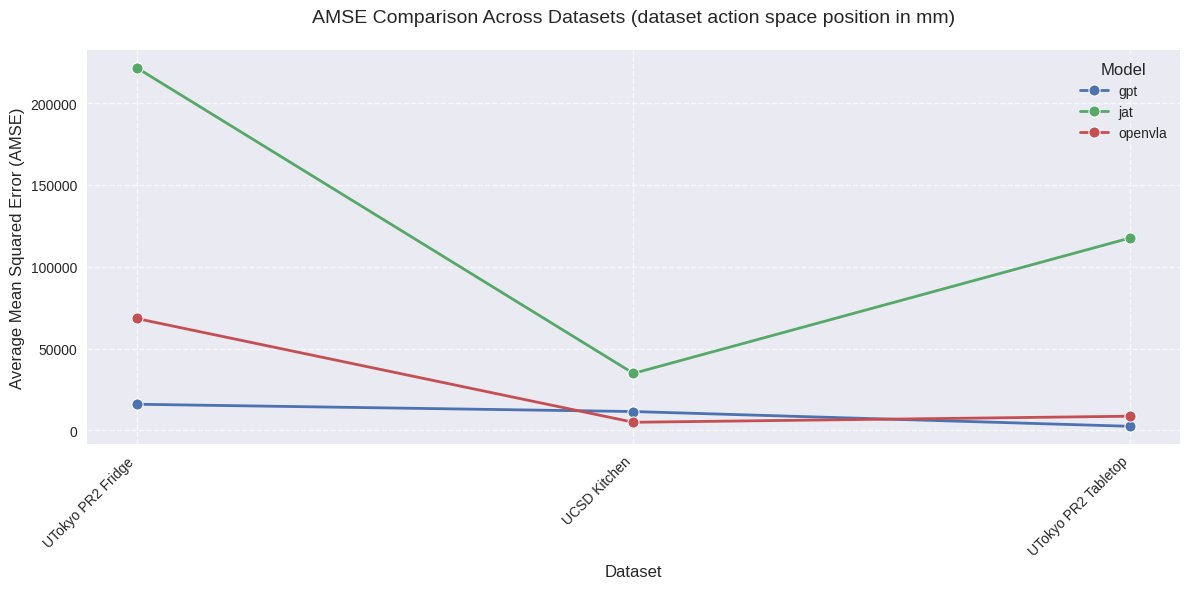}
	\caption{AMSE Across Datasets with Action Space Unit in Millimeter}
	\label{fig:fig5}
\end{figure}

\end{document}